\def\year{2018}\relax
\documentclass[letterpaper]{article} 
\usepackage{aaai18}  
\usepackage{times}  
\usepackage{helvet}  
\usepackage{courier}  
\usepackage{url}  
\usepackage{graphicx}  
\usepackage{comment}
\usepackage{latexsym}
\usepackage{amsmath}
\usepackage{amsfonts}
\usepackage{amssymb}
\usepackage{dsfont}
\usepackage[english]{babel}
\usepackage{color}
\usepackage{booktabs}
\usepackage{multicol}
\usepackage[dvipsnames]{xcolor}
\usepackage{algorithm,algcompatible,amsmath}

\begin{document}
%

\title{R$^3$: Reinforced Ranker-Reader for Open-Domain Question Answering}

\author{ Shuohang Wang$^1$\thanks{~~Word done when the author was at IBM.}, Mo Yu$^2$, Xiaoxiao Guo$^2$, Zhiguo Wang$^2$, Tim Klinger$^2$, Wei Zhang$^2$\\
{ \bf \Large Shiyu Chang$^2$, Gerald Tesauro$^2$, Bowen Zhou$^3$, and Jing Jiang$^1$} \\
$^1${School of Information System, Singapore Management University}\\
$^2${AI Foundations - Learning, IBM Research AI. Yorktown Heights NY, USA}\\
$^3${JD.COM. Beijing, China}\\
{\texttt {shwang.2014@smu.edu.sg,
 yum@us.ibm.com,
xiaoxiao.guo@ibm.com}}
}
 
\maketitle
\begin{abstract}
In recent years researchers have achieved considerable success applying neural network methods to question answering (QA).  These approaches have achieved state of the art results in simplified closed-domain settings\footnote{In the QA community, ``openness'' can be interpreted as referring either to the scope of question topics or to the breadth and generality of the knowledge source used to answer each question. Following \cite{chen2017reading} we adopt the latter definition.} such as the SQuAD~\cite{rajpurkar2016squad} dataset, which provides a pre-selected passage, from which the answer to a given question may be extracted.  More recently, researchers have begun to tackle {\it open-domain QA}, in which the model is given a question and access to a large corpus (e.g., wikipedia) instead of a pre-selected passage ~\cite{chen2017reading}.
This setting is more complex as it requires large-scale search for relevant passages by an information retrieval 
component, combined with a reading comprehension 
model that
``reads'' the passages to generate an answer to the question.
Performance in this 
setting 
lags well behind closed-domain performance.

In this paper, we present a novel open-domain QA system called {\it{Reinforced Ranker-Reader}} 
$(R^3)$, based on two algorithmic innovations.  
First, we propose a new pipeline for open-domain QA with a \textit{Ranker} component, which learns to rank retrieved passages in terms of likelihood of extracting the ground-truth answer to a given question.
Second, we propose a novel method that jointly trains the \textit{Ranker} along with an answer-extraction \textit{Reader} model, based on reinforcement learning.
We report extensive experimental results showing that our method significantly improves on the state of the art for multiple open-domain QA datasets.~\footnote{Code:  \url{https://github.com/shuohangwang/mprc}. }
\end{abstract}

\section{Introduction}
Open-domain question answering (QA) is a key challenge in natural language processing. A successful open-domain QA system must be able to effectively retrieve and comprehend one or more knowledge sources to infer a correct answer. Knowledge sources can be knowledge bases \cite{berant2013semantic,yu2017improved} or structured or unstructured text passages \cite{ferrucci2010building,baudivs2015modeling}.

Recent deep learning-based research has focused on open-domain QA based on large text corpora such as Wikipedia, applying 
information retrieval (IR) to select passages and reading comprehension (RC) to extract answer phrases~\cite{chen2017reading,dhingra2017quasar}. 
These methods, which we call \emph{Search-and-Reading QA} (SR-QA), are simple yet powerful for open-domain QA.  Dividing the pipeline into IR and RC stages leverages an enormous body of research in both IR and RC, including recent successes in RC via neural network techniques
~\cite{wang2016machine,wang2016multi,xiong2016dynamic,wang2017gated}.

\begin{table}[]
\centering
\small
\begin{tabular}{ll}
Q: & \multicolumn{1}{p{7cm}}{What is the largest island in the Philippines?}                                                      \\
A: & \multicolumn{1}{p{7cm}}{\textbf{Luzon}}\\
P1 & \multicolumn{1}{p{7cm}}{Mindanao is the second largest and easternmost island in the Philippines.}                               \\
P2 & \multicolumn{1}{p{7cm}}{As an island, \textbf{Luzon} is the Philippine's largest at 104,688 square kilometers, and is also the world's 17th largest island.}                                       \\
P3 & \multicolumn{1}{p{7cm}}{Manila, located on east central \textbf{Luzon} Island, is the national capital and largest city. }                                       \\
\end{tabular}
\normalsize
\caption{An open-domain QA training example. Q: question, A: answer, P: passages retrieved by an IR model and ordered by IR score. }
\label{example}
\end{table}
The main difference between training SR-QA and standard RC models is in the passages used for training. In standard RC model training, passages are manually selected to guarantee that ground-truth answers are contained and annotated within the passage~\cite{rajpurkar2016squad}.\footnote{This forms a closed-domain QA by our adopted definition where the domain consists of the given passage only.}
By contrast, in SR-QA approaches~\cite{chen2017reading,dhingra2017quasar}, the model is given only QA-pairs and uses an IR component to retrieve passages similar to the question from a large corpus. 
Depending on the quality of the IR component, retrieved passages may not contain or entail the correct answer, making RC training more difficult.
Table 1 shows an example which illustrates the difficulty. This ordering was produced by an off-the-shelf IR engine using the BM25 algorithm. The correct answer is contained in passage P2. The top passage (P1), despite being ranked highest by the IR engine, is ineffective for answering the question, since it fails to capture the semantic distinction between ``largest'' and ``second largest''. Passage P3 contains the answer text (``Luzon'') but does not semantically entail the correct answer (``Luzon is the largest island in the Philippines''). Training on passages such as P1 and P3 can degrade performance of the RC component.\footnote{Passage ranking models for non-factoid QA~\cite{wang2007jeopardy,yang2015wikiqa} are able to learn to rank these passages; but these models are trained using human annotated answer labels, which are not available here.}  

In this paper we propose a new approach 
which explicitly separates the tasks of predicting the likelihood that a passage 
provides 
the answer, and reading those passages to 
extract correct answers.  Specifically we propose an end-to-end framework consisting of two components: a \textit{Ranker} and a \textit{Reader} (i.e. RC model).  The Ranker selects the passage most likely to entail the answer and passes it to the Reader, which reads and extracts from that passage. The Reader is trained using SGD/backprop to maximize the likelihood of the span containing the correct answer (if one exists). The Ranker is trained using REINFORCE~\cite{Williams1992} with a reward determined by how well the Reader extracts answers from the top-ranked passages. This optimizes the Ranker with an objective determined by end-performance on answer prediction, which provides a strong signal to distinguish passages lexically similar to but semantically different from the question.

We discuss the Ranker-Reader model in detail below but briefly, the Ranker and Reader are implemented as variants of
Match-LSTM models~\cite{wang2015learning:NAACL2016}.  These models were originally designed for solving the text entailment problem. For this task, different non-linear layers are added for selecting the passages or predicting the start and end positions of the answer in the passage. 

We evaluate our model on five different datasets and achieve state-of-the-art results on four of the them.  Our results also show the merits of employing a separate REINFORCE-trained ranking component over several challenging fully supervised baselines.

\section{Framework}

\noindent\textbf{Problem Definition}\quad
We assume that we have available a factoid question $\mathbf{q}$ to be answered and a set of passages which may contain the ground-truth answer $\mathbf{a}^g$.  Those passages\footnote{In this paper we use sentence-level index thus each passage is an individual sentence. See the experimental setting.} are the top $N$ retrieved from a corpus by an IR model supplied with the question, for $N$ a hyper-parameter.
During training we are given only the $(\mathbf{q}, \mathbf{a}^g)$ pairs, together with an IR model with index built on an open-domain corpus.\\

\noindent\textbf{Framework Overview} \quad
An overview of the Ranker-Reader model is shown in Figure~\ref{fig:model}. It shows two key components: a \textbf{Ranker}, which selects passages from which an answer can be extracted, and a \textbf{Reader} which extracts answers from supplied passages.  Both the Ranker and Reader 
compare the question to each of the passages to generate passage representations based on how well they match the question. The Ranker uses these ``matched'' representations to select a single passage which is 
most likely to contain the answer.  The selected passage is then processed by the Reader to extract an answer sequence.  
We train the reader using SGD/backprop and produce a reward to train the Ranker via REINFORCE. 

\section{$\textbf{R}^{3}$: Reinforced Ranker-Reader}
\label{sec:method}

\begin{figure*}[t]
\centering
\includegraphics[width=7in]{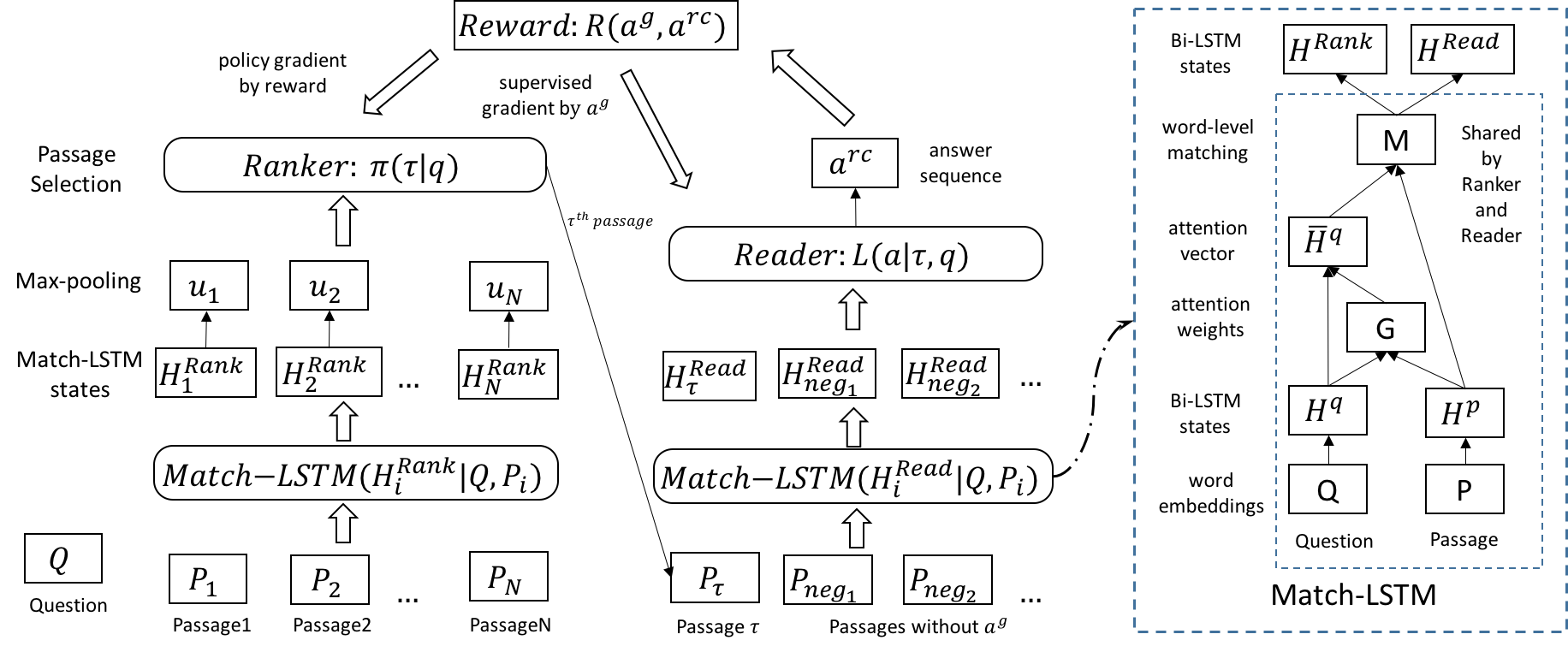}

\caption{Overview of training our model, comprising a Ranker and a Reader based on Match-LSTM as shown on the right side. The Ranker selects a passage $\tau$ and the Reader predicts the start and end positions of the answer in $\tau$. The reward for the Ranker depends on similarity of the extracted answer with the ground-truth answer $\mathbf{a}^g$. To accelerate Reader convergence,
we also sample several negative passages without ground-truth answer.}
\label{fig:model}
\end{figure*}

In this section, we first review the Match-LSTM~\cite{wang2015learning:NAACL2016} which 
provides input for both the Reader and Ranker. We then detail the Reader and Ranker components, and the procedure for joint training, including the objective function used for RL training.

\paragraph{Passage Representation Using Match-LSTM}
To effectively rank and read passages they must be matched to the question.  This comparison is performed with a Match-LSTM, a state-of-the-art model for text entailment, shown on the right in~Figure~\ref{fig:model}. Match-LSTMs use an attention mechanism to compute word similarities between the passage and question sequences.  These are first encoded as matrices $\mathbf{Q}$ and $\mathbf{P}$, respectively, by a Bidirectional LSTM (BiLSTM) with hidden dimension $l$.  With $Q$ words in question $\mathbf{Q}$ and $P$ words in passage $\mathbf{P}$ we can write:

\begin{equation}
\begin{matrix}
\mathbf{H}^{\text{p}} = \text{BiLSTM} (\mathbf{P}), & \mathbf{H}^{\text{q}} =\text{BiLSTM} (\mathbf{Q}),
\end{matrix}
\label{eqn:preprocess}
\end{equation}
where $\mathbf{H}^{\text{p}} \in \mathbb{R}^{l\times P}$ and $ \mathbf{H}^{\text{q}}\in \mathbb{R}^{l\times Q}$ are the hidden states for the passage and the question. 
In order to improve computational efficiency without degrading performance, we simplify the attention mechanism of the original Match-LSTM by computing the attention weights $\mathbf{G}$ as follows:

\begin{eqnarray}
\nonumber
\mathbf{G} & = & \text{SoftMax}\left( (\mathbf{W}^{\text{g}} \mathbf{H}^{\text{q}}+\mathbf{b}^\text{g}\otimes \mathbf{e}_Q)^\text{T} \mathbf{H}^{\text{p}} \right)
\label{eqn:alpha}
\end{eqnarray}
where $\mathbf{W}^{\text{g}}\in \mathbb{R}^{l\times l}$ and $\mathbf{b}^\text{g} \in \mathbb{R}^{l}$ are the learnable parameters. The outer product $(\cdot \otimes \mathbf{e}_Q)$ repeats the column vector $\mathbf{b}^{\text{g}}$ $Q$ times to form an $l \times l$ matrix. The $i$-th column of $\mathbf{G}\in \mathbb{R}^{Q\times P}$ represents the normalized attention weights over all the question words for the $i$-th word in passage.

We can use this attention matrix $G$ to form representations of the question for each word in passage:

\begin{equation}
\overline{\mathbf{H}}^{\text{q}} = \mathbf{H}^{\text{q}}\mathbf{G}
\end{equation}

Next, we produce the word matching representations $\mathbf{M}\in \mathbb{R}^{2l\times P}$ using  $\mathbf{H}^\text{p}$ and $\overline{\mathbf{H}}^\text{q}$ as follows:

\begin{eqnarray}
\mathbf{M} & = & \text{ReLU}\left(  \mathbf{W}^\text{m} \begin{bmatrix}
\mathbf{H}^{\text{p}}
\\ 
\overline{\mathbf{H}}^{\text{q}}
\\ 
 \mathbf{H}^{\text{p}} \bigodot \overline{\mathbf{H}}^{\text{q}}
\\ 
\mathbf{H}^{\text{p}} - \overline{\mathbf{H}}^{\text{q}}
\end{bmatrix} \right),
\label{eqn:match}
\end{eqnarray}
where $\mathbf{W}^\text{m}\in \mathbb{R}^{2l\times 4l}$ are learnable parameters; $\begin{bmatrix}
\cdot \\ \cdot \end{bmatrix}$ is the column concatenation of matrices; Element-wise operations $(\cdot \bigodot \cdot )$ and $(\cdot - \cdot )$ are also used to represent word-level matching~\cite{wang2016compare,Chen:2017:ACL}.

Finally, we aggregate the word matching representations through another bi-directional LSTM:

\begin{equation}
\mathbf{H}^{\text{m}} = \text{BiLSTM} (\mathbf{M}),
\label{eqn:matchlstm}
\end{equation}
where $\mathbf{H}^{\text{m}} \in \mathbb{R}^{l\times P}$ is the sequence matching representation between a passage and a question.

To produce the input for the Ranker and Reader described next, we apply Match-LSTMs to the question and each of the passages.  To reduce model complexity, the Ranker and Reader share the same $\mathbf{M}$ but have separate parameters for the aggregation stage shown in Eqn.(\ref{eqn:matchlstm}), resulting different $\mathbf{H}^{\text{m}}$, denoted as $\mathbf{H}^{\text{Rank}}$ and $\mathbf{H}^{\text{Read}}$ respectively.

\paragraph{Ranker}
Our Ranker selects passages for reading by the Reader.
We train the Ranker using reinforcement learning, to output a policy or probability distribution over passages. 
First, we create a fixed-size vector  representation for each passage from the matching representations  $\mathbf{H}^{\text{Rank}}_{\text{i}}$, $\text{i} \in [1,N]$, using a standard max pooling operation.  The result $\textbf{u}_{\text{i}}$ is a representation of the $\text{i}$-th passage. We then concatenate the individual passage representations and apply a non-linear transformation followed by a normalization to compute the passage probabilities $\gamma$.  Specifically:
\begin{eqnarray}
\nonumber
\textbf{u}_i &=& \text{MaxPooling}(\mathbf{H}^{\text{Rank}}_\text{i}), \\
\nonumber
\mathbf{C} &=& \text{Tanh}\left( \mathbf{W}^\text{c}[\textbf{u}_1;\textbf{u}_2;...;\textbf{u}_N] + \mathbf{b}^\text{c} \otimes \mathbf{e}_{N} \right ), \\
\gamma &=& \text{Softmax}(\mathbf{w}^c \mathbf{C}),
\label{eqn:gamma}
\end{eqnarray}
where $\mathbf{W}^\text{c}\in \mathbb{R}^{l\times l}$ and $\mathbf{b}^\text{c},\mathbf{w}^\text{c}\in \mathbb{R}^{l}$ are the parameters to optimize; $\textbf{u}_i \in \mathbb{R}^{l}$ represents how the $i^{\text{th}}$ passage matches the question; $\mathbf{C}\in \mathbb{R}^{l\times N}$ is a non-linear transformation of passage representations; and $\gamma\in \mathbb{R}^N$ is a vector of the predicted probabilities that each passage entails the answer.

The action policy is then defined as follows:

\begin{equation}
\pi (\tau | \mathbf{q}; \theta^{r}) = \gamma _\tau 
\label{eqn:pi}
\end{equation}
where $\gamma _\tau $ is the probability of selecting passage $\tau$, computed in Eqn.(\ref{eqn:gamma}); $\theta^r$ represents parameters to learn. In the rest of the paper we denote the policy $\pi(\tau|q) = \pi (\tau | q; \theta^{r})$ for simplicity. In this way, the action is to sample a passage according to its policy $\pi(\tau|q)$ as the input of Reader.

\paragraph{Reader}
Our Reader extracts an answer span from the passage $\tau$ selected by the Ranker. As in previous work~\cite{wang2016machine,xiong2016dynamic,seo2016bidirectional,wang2017gated}, the Reader is used to predict the start and end positions of the answer phrase in the passage.

First we process the output of Match-LSTMs on all the passages to produce the probability of the start position of the answer span $\mathbf{\beta}^{\text{s}}$:
\begin{eqnarray}
\nonumber
\mathbf{F}^\text{s} &=& \text{Tanh}\left( \mathbf{W}^\text{s}[\mathbf{H}^{\text{Read}}_\tau;\mathbf{H}^{\text{Read}}_{\text{neg}_1};...;\mathbf{H}^{\text{Read}}_{\text{neg}_n}] + \mathbf{b}^\text{s} \otimes \mathbf{e}_{V} \right ), \\
\mathbf{\beta}^\text{s} &=& \text{Softmax}\left( \mathbf{w}^\text{s} \mathbf{F}^\text{s} \right ),
\label{eqn:span}
\end{eqnarray}
where $neg_n$ is the id of a sampled passage not containing ground-truth answer during training;
$V$ is the total number of words in these passages; $\mathbf{e}_{V}$ is thus a $V$-dimension vector with ones; $[\cdot ; \cdot]$ is the column concatenation operation; $\mathbf{W}^\text{s}\in \mathbb{R}^{l\times l}$ and $\mathbf{b}^\text{s},\mathbf{w}^\text{s}\in \mathbb{R}^{l}$ are the parameters to optimize; $\mathbf{\beta}^\text{s}\in \mathbb{R}^{V}$ is the probability of the start point of the span.

We similarly compute the probability of the ending position, $\mathbf{\beta}^\text{e}\in \mathbb{R}^{V}$, using separate parameters $\mathbf{W}^\text{e}, \mathbf{b}^\text{e}$ and $\mathbf{w}^\text{e}$.
The loss function can then be expressed as follows:
\begin{equation}
L(\mathbf{a}^g|\tau, \mathbf{q})=-\text{log} (\mathbf{\beta}^\text{s}_{a^s_\tau}) -
\text{log} (\mathbf{\beta}^\text{e}_{a^e_\tau}) , 
\label{eqn:rc_obj}
\end{equation}
where $\mathbf{a}^g$ is the ground-truth answer; $\tau$ is sampled according to Eqn.(\ref{eqn:pi}), and during training, we keep sampling until passage $\tau$ contains $\mathbf{a}^g$; $\mathbf{\beta}^\text{s}_{a^s_\tau}$ and $\mathbf{\beta}^\text{e}_{a^e_\tau}$ represent the probability of the start and end positions of $\mathbf{a}^g$ in passage $\tau$. 

\paragraph{Training}

We adopt joint training of Ranker and Reader as shown in Algorithm 1.  Since the Ranker makes a hard selection of the passage, it is trained using the REINFORCE algorithm. The Reader is trained using standard
SGD/backprop.

\begin{algorithm}
	\label{algorithm:alg1}
    \caption{Reinforced Ranker-Reader ($\text{R}^{3}$)}
  \begin{algorithmic}[1]
    \State \textbf{Input: }$\mathbf{a}^{g}$, $\mathbf{q}$, passages from IR
    \State \textbf{Output:} $\Theta$ 
    \State \textbf{Initialize:}
     $\Theta \gets$ pre-trained $\Theta$ with a baseline method\footnotemark
    \FOR{ each $\mathbf{q}$ in dataset}
      \STATE For question $\mathbf{q}$, sample $K$ passages from the top $N$ passages retrieved by IR model for training. \footnotemark
      \STATE Randomly sample a positive passage $\tau \sim \pi (\tau | \mathbf{q})$
      \STATE Extract the answer $\mathbf{a}^{rc}$ through RC model
      \STATE Get reward $r$ according to $R(\mathbf{a}^g,\mathbf{a}^{rc}|\tau)$.
      \STATE {Updating Ranker (ranking model) through policy gradient $r\frac{\partial}{\partial \Theta}\log(\pi(\tau|\mathbf{q}))$ 
      \STATE Updating Reader (RC model) through supervised gradient $\frac{\partial}{\partial \Theta} L (\mathbf{a}^g | \tau, \mathbf{q})$ }
	\ENDFOR
  \end{algorithmic}
\end{algorithm}
 \addtocounter{footnote}{-2} 
 \stepcounter{footnote}
 \footnotetext{Baseline method $\text{SR}^{2}$, described in Experimental Settings.}
 \stepcounter{footnote}
\footnotetext{For computational efficency, we sample 10 passages during training, and make sure there are at least 2 negative passages and as many positive passages as possible.}

Our training objective is to minimize the following loss function

\begin{equation}
J(\Theta)=
-\mathbb{E}_{\tau  \sim \pi (\tau | \mathbf{q} )} \left[ L(\mathbf{a}^g|\tau, \mathbf{q})\right],
\end{equation}
where $L$ is the loss of the Reader defined in Eqn. (\ref{eqn:rc_obj}); $\pi (\tau | \mathbf{q})$ is the action policy 
defined in Eqn.(\ref{eqn:pi});  and $\Theta$ are parameters to be learned.
During training, action sampling 
is limited solely to passages containing the ground-truth answer, to guarantee Reader updating (line 10 in Algorithm 1) based on the sampled passages with supervised gradients. 
The gradient of $J(\Theta)$ with respect to $\Theta$ is:

\begin{equation}
\small
\begin{split}
\nabla_{\Theta}  J(\Theta) 
&= -\nabla_{\Theta} \sum_{\tau}  \pi(\tau|\mathbf{q}) L (\mathbf{a}^g | \tau, \mathbf{q}) \\
&=  - \sum_{\tau} \left( L(\mathbf{a}^g | \tau, \mathbf{q}) \nabla_{\Theta} \pi(\tau|\mathbf{q})  + \pi(\tau|\mathbf{q}) \nabla_{\Theta} L (\mathbf{a}^g | \tau, \mathbf{q}) \right) \\
&= - \mathbb{E}_{\tau  \sim \pi(\tau|\mathbf{q})}\left[ L (\mathbf{a}^g | \tau, \mathbf{q}) \nabla_{\Theta}\log(\pi(\tau|\mathbf{q})) \right. \\
&\quad\quad\quad + \left. \nabla_{\Theta} L (\mathbf{a}^g | \tau, \mathbf{q}) \right]\\
&\approx - \mathbb{E}_{\tau  \sim \pi(\tau|\mathbf{q})}\left[ R (\mathbf{a}^g,\mathbf{a}^{rc} | \tau) \nabla_{\Theta} \log(\pi(\tau|\mathbf{q})) \right. \\
&\quad \quad \quad + \left. \nabla_{\Theta} L (\mathbf{a}^g | \tau, \mathbf{q}) \right]
\label{eqn:rl_obj}
\end{split}
\normalsize
\end{equation}

So in training, we first sample a passage $\tau$ according to the policy $\pi (\tau | \mathbf{q})$. Then the Reader updates its parameters given the passage $\tau$ using standard Backprop and the ranker updates its parameters via policy gradient using $L(a|\tau,\mathbf{q})$ as rewards. However, $L(a|\tau,\mathbf{q})$ is not bounded and introduces a large variance in gradients (similar to what was reported in ~\citeauthor{mnih2014recurrent}~\citeyear{mnih2014recurrent}). To address this, we replace $L(a|\tau,\mathbf{q})$ with a bounded reward $R(\mathbf{a}^g,\mathbf{a}^{rc}|\tau)$, which captures how well the answer extracted by the Reader matches the ground-truth answer.  Specifically:

\begin{equation}
\small
R(\mathbf{a}^g,\mathbf{a}^{rc}| \tau) =\left\{\begin{matrix}
2, & \text{if} \; \mathbf{a}^g == \mathbf{a}^{rc}\\ 
f1(\mathbf{a}^{g},\mathbf{a}^{rc}), & \text{else if}\; \mathbf{a}^{g} \cap \mathbf{a}^{rc} !=\o  \\ 
-1, & \text{else}
\end{matrix}\right.
\end{equation}
where $\mathbf{a}^g$ is the ground-truth answer; $\mathbf{a}^{rc}$ is the answer extracted by Reader; $f1(\cdot ,\cdot )\in [0,1]$ 
computes
word-level F1 score between two sequences. F1 is used as reward when 
$\mathbf{a}^g$ and $\mathbf{a}^{rc}$ share some words but do not exactly match. We give a larger reward of 2 for exact match, and -1 reward for no overlap. \\

\noindent\textbf{Prediction} \quad
During testing, we combine the Ranker and Reader for answer extraction as follows: 
\begin{equation}
\Pr(\mathbf{a},\tau) = \Pr(\mathbf{a}|\tau)\Pr(\tau) =
e^{-L(\mathbf{a}|\tau,\mathbf{q}) } \pi(\tau|\mathbf{q}),
\label{eqn:pred_prob}
\end{equation}
where $\Pr(\mathbf{a},\tau)$ is the probability of extracting the answer~$\textbf{a}$ from passage $\tau$. We select the answer with the largest $\Pr(\mathbf{a},\tau)$ as the final prediction.

\section{Experimental Settings}
To evaluate our model we have chosen five challenging datasets under the open-domain QA setting  and three public baseline models.

\subsection{Datasets}
We experiment with five different datasets whose statistics are shown in Table~\ref{datasets}.

\textbf{Quasar-T} is a dataset for
SR-QA, with
question-answer pairs from various internet sources. 
Each question is compared to 100 sentence-level candidate passages, retrieved by their IR model from the ClueWeb09 data source, to extract the answer. 

The other four datasets we consider are: \textbf{SQuAD}, the Stanford QA dataset, from which we take only the question-answer pairs and discard the passages to form an open-domain QA setting (denoted as \textbf{SQuAD$_{\textrm{OPEN}}$});
\textbf{WikiMovies} which contains movie-related questions from the OMDb and MovieLens databases and where the questions can be answered using Wikipedia pages; \textbf{CuratedTREC}, based on TREC~\cite{voorhees2000building} and designed for open-domain QA; and \textbf{WebQuestion} which is designed for knowledge-base QA with answers restricted to Freebase entities.
For these four datasets under the open-domain QA setting, no candidate passages are provided so we build a similar sentence-level Search Index based on English Wikipedia,
following~\citeauthor{chen2017reading}~\citeyear{chen2017reading}'s work.
To provide a small yet sufficient search space for our model, we employ a traditional 
IR method to retrieve relevant passages from the whole of Wikipedia. We use the 2016-12-21 dump of English Wikipedia as our sole knowledge source, and build an inverted index with Lucene\footnote{\url{https://lucene.apache.org/}}. We then take each input question as a query to search for top-200 articles, rank them with BM25, and split them into sentences.  The sentences are then ranked by TF-IDF and the top-200 sentences for each question retained.

\begin{table}[]
\centering
\small
\begin{tabular}{lcccc}
\toprule
            & \#q(train) & \#q(test)  & \#p(train) & \#p(test)\\
\midrule
Quasar-T    & 28496 & 3000  & 14.8 / 100 & 1.9 / 50 \\
SQuAD$_\text{OPEN}$       & 82271 & 10570 & 35.1 / 200 & 2.3 / 50 \\
WikiMovies  & 93935 & 9,952 & 68.5 / 200 & 1.8 / 50 \\
CuratedTREC & 1204  & 694   & 14.6 / 200 & 4.8 / 50 \\
WebQuestion & 3272  & 2,032 & 57.2 / 200 & 4.1 / 50 \\
\bottomrule
\normalsize
\end{tabular}
\caption{Statistics of the datasets. \#q represents the number of questions. For the training dataset, we ignore the questions without any answer in all the retrieved passages. In the special case that there's only one answer for the question, during training, we combine the question with the answer as the query to improve IR recall.  Otherwise we use only the question. \#p represents the number of passages and 14.8~/~100 means there are 14.8 passages containing the answer on average out of the 100 passages. We use top50 passages retrieved by the IR model for testing. }
\label{datasets}
\end{table}

\subsection{Baselines}
We consider three public baseline models\footnote{We only compare to the results from the public papers.}: GA~\cite{dhingra2016gated,dhingra2017quasar}, a gated-attention reader for text comprehension; BiDAF~\cite{seo2016bidirectional}, a reader with bidirectional attention flow for machine comprehension; and DrQA~\cite{chen2017reading}, a document reader for question answering. We also compare our model $\text{R}^3$ with two internal baselines:\\

\noindent \textbf{Single Reader (SR)}\quad 
This model is trained in the same way as~\citeauthor{chen2017reading}~\citeyear{chen2017reading} and~\citeauthor{dhingra2017quasar}~\citeyear{dhingra2017quasar}. We find all the answer spans that exactly match the ground-truth answers from the retrieved passages and train the  Reader using the objective of Eqn.(\ref{eqn:rc_obj}).  Here $\tau$ is randomly sampled from $[1,N]$ instead of using Eqn.(\ref{eqn:pi}).\\

\noindent \textbf{Simple Ranker-Reader ($\text{SR}^2$)}\quad
This Ranker-Reader model is trained by combining the two different objective functions for the Single Reader and the Ranker models together. In order to train the Ranker, we treat all the passages that contain the ground-truth answer as positive cases and use the following for the Ranker loss:

\begin{equation}
\sum_{n=1}^{N} y_n \left( \log(y_n) - \log(\gamma_n) \right),
\label{eqn:rank_obj_ideal}
\end{equation}
which is the KL divergence between $\mathbf{\gamma}$ computed through Eqn.(\ref{eqn:gamma}) and a probability vector $\mathbf{y}$, where $y_i = {1}/{N_p}$ when the passage $i$ contains the ground-truth answer, and $y_i = {0}/{N_p}$ otherwise.
$N_p$ is the total number of passages which contain the ground-truth answer in the top-$N$ passage list.
\subsection{Implementation Details}
In order to increase the likelihood that question-related context will be contained in the retrieved passages for the training dataset, if the answer is unique, we combine the question with the answer to form the query for
information retrieval. For the testing dataset, we use only the question as a query and collect the top 50 passages for answer extraction.

During training, our $\text{R}^3$ model is first initialized by pre-training the model using the Simple Ranker-Reader ($\text{R}^2$), to encourage convergence.
As discussed earlier, the pre-processing and matching layers, Eqn.(\ref{eqn:preprocess}-\ref{eqn:match}), are shared by both Ranker and Reader. The number of LSTM layers in Eqn.(\ref{eqn:matchlstm}) is set to 3 for the Reader and 1 for the Ranker.

Our model is optimized using Adamax~\cite{kingma2014adam:iclr2015}. We use fixed GloVe~\cite{glove:emnlp2014} 
word embeddings.
We set $l$ to 300, batch size to 30, learning rate to 0.002 and tune the dropout probability.

\begin{table*}[t]
\centering
\small
\begin{tabular}{lcccccccccc}
\toprule
                  & \multicolumn{2}{c}{\bf Quasar-T} & \multicolumn{2}{c}{\bf SQuAD$_{\textrm{OPEN}}$} & \multicolumn{2}{c}{\bf WikiMovies}  & \multicolumn{2}{c}{\bf CuratedTREC} & \multicolumn{2}{c}{\bf WebQuestions}\\
                  & \bf F1            & \bf EM            & \bf F1            & \bf EM            & \bf F1               & \bf EM               & \bf F1               & \bf EM               & \bf F1               & \bf EM              \\
\midrule
GA~\cite{dhingra2016gated}               & 26.4          & 26.4          & -             & -             & -                & -                & -                & -                & -                & -               \\
BiDAF~\cite{seo2016bidirectional}           & 28.5          & 25.9          & -             & -             & -                & -                & -                & -                & -                & -               \\

DrQA~\cite{chen2017reading}             & -             & -             & -             & 28.4  & -                & 34.3        & -                & 25.7             & -                & $\mathbf{19.5}$                         \\
\midrule
Single Reader (SR)    & $\text{38.5}^{.2}$          & $\text{31.5}^{.2}$          & $\text{35.4}^{.2}$         & $\text{26.9}^{.2}$   & $\text{38.8}^{.1}$           & $\text{37.7}^{.1}$       & $\text{33.6}^{.6}$            & $\text{27.4}^{.4}$           & $\text{22.0}^{.2}$            & $\text{15.2}^{.3}$                         \\
Simple Ranker-Reader ($\text{SR}^2$)       & ${38.8}^{.2} $         & $\text{31.9}^{.2}$          & $\text{35.8}^{.2}$          & $\text{27.2}^{.2}$         & $\text{39.3}^{.1}$              & $\text{38.1}^{.1}$  & $\text{33.4}^{.6}$            & $\text{27.7 }^{.5}$             & $\text{22.5}^{.3} $            & ${15.6 }^{.4}$                          \\
Reinforced Ranker-Reader ($\text{R}^3$)  & $\mathbf{40.9}^{.3}$ & $\mathbf{34.2}^{.3}$ & $\mathbf{37.5}^{.2}$ & $\mathbf{29.1}^{.2}$ & $\mathbf{39.9}^{.1}$    & $\mathbf{38.8}^{.1}$ & $\mathbf{34.3}^{.6}$    & $\mathbf{28.4}^{.6}$    & $\mathbf{24.6}^{.3}$    & $\text{17.1}^{.3}$      \\
\hline
\hline
DrQA-MTL~\cite{chen2017reading}               & -             & -             & -             & 29.8     & -                & 36.5     & -                & 25.4             & -                & 20.7                         \\
YodaQA~\cite{baudivs2015modeling}           & -             & -             & -             & -          & -           & -                & -      & 31.3             & -                & 39.8                        \\
\bottomrule
\end{tabular}
\normalsize
\caption{Open-domain question answering results. The results show the average of 5 runs, with standard error in the superscript.  The CuratedTREC and WebQuestions models are initialized by training on SQuAD$_{\textrm{OPEN}}$ first. On the bottom, YodaQA and DrQA-MTL use additional resources (usage of KB for the former, and multiple training datasets for the latter), so are not a true apple-to-apple comparison to the other methods. EM: Exact Match.  }
\label{tab:results}
\end{table*}

\section{Results and Analysis}
In this section, we will show the performance of different models on five QA datasets and offer further analysis. 
\subsection{Overall Results}
Our results are shown in Table~\ref{tab:results}. We use F1 score and Exact Match (EM) evaluation metrics\footnote{Evaluation tooling is from SQuAD~\cite{rajpurkar2016squad}.}.  We first observe that on Quasar-T, the Single Reader can exceed state-of-the-art performance. Moreover, unlike DrQA, our models are all trained using distant supervision and, without pre-training on the original SQuAD dataset\footnote{The performance of our Single Reader model on the original SQuAD dev set is F1 77.0, EM 67.6 which is close to the BiDAF model, F1 77.3, EM 67.7 and DrQA model, F1 78.8, EM 69.5.}, our Single Reader model still achieves better performance on the WikiMovie and CuratedTREC datasets. 

Next we observe that the Reinforced Ranker-Reader~($\text{R}^3$) achieves the best performance on the Quasar-T, WikiMovies, and CuratedTREC datests and achieves significantly better performance than our internal baseline model Simple Ranker-Reader ($\text{SR}^2$) on all datasets except CuratedTREC.
These results demonstrate the effectiveness of using RL to jointly train the Ranker and Reader both as compared to competing approaches and the non-RL Ranker-Reader baseline.

\begin{table}[t]
\centering
\begin{tabular}{lcc}
\toprule
                  & F1            & EM \\
                  \midrule
Single Reader (SR) & 38.3 & 31.4 \\
SR + Ranker (from $\text{SR}^2$)  & 38.9 & 31.8 \\
SR + Ranker (from $\text{R}^3$)  & \textbf{40.0} & \textbf{33.1}  \\ 
\midrule
$\text{SR}^2$  & 38.7 & 31.9 \\
$\text{R}^3$  & {40.8} & {34.1}  \\ 
\bottomrule
                  \end{tabular}
\caption{Effects of rankers from $\text{SR}^2$ and $\text{R}^3$ (on Quasar-T test dataset). Here we use the same single reader model (SR) as the reader, combined with two different rankers.
The performance of the two runs of $\text{SR}^2$ and $\text{R}^3$ (that provide the rankers) is listed at bottom.
}
\label{tab:analysis1}
\end{table}

\begin{table}[h]
\centering
\begin{tabular}{lccc}
\toprule
                 & TOP-k & F1            & EM \\
                  \midrule
Single Reader (SR) & 1 & 38.3 & 31.4 \\
Single Reader (SR) & 3 & 51.7 & 43.7 \\
Single Reader (SR) & 5 & 58.7 & 49.2 \\
SR + Ranker (from $\text{R}^3$) & 1  & \textbf{40.0} & \textbf{33.1} \\
\bottomrule
                  \end{tabular}
\caption{Potential improvement on QA performance by improving the ranker. The performance is based on the Quasar-T test dataset.  
The \textbf{TOP-3/5} performance is used to evaluate the further potential improvement by improving rankers (see the ``Potential Improvement'' section).
}
\label{tab:analysis_upper}

\end{table}

\begin{table*}[]
\centering
\small
\begin{tabular}{lll}
\toprule
Q & \multicolumn{2}{l}{Apart from man what is New Zealand 's only native mammals}  \\
A  & \multicolumn{2}{l}{\textbf{bats}} \\
\midrule
         & \multicolumn{1}{c}{Reinforced Ranker-Reader ($\text{R}^3$)}                           & \multicolumn{1}{c}{Simple Ranker-Reader ($\text{SR}^2$)}                                                                                         \\
P1 & \multicolumn{1}{p{8cm}}{New Zealand has no native land mammals apart from some rare \textbf{bats} . 
} &\multicolumn{1}{p{8cm}}{ New Zealand 's native species were \textbf{sitting ducks} ! 
} \\
P2 & \multicolumn{1}{p{8cm}}{New Zealand 's native species were sitting \textbf{ducks} !   
}  & 
\multicolumn{1}{p{8cm}}{1080 is a commonly used pesticide since it is very effective on mammals and New Zealand has no native land mammals apart from two species of \textbf{bat} .                          
} \\
P3 & \multicolumn{1}{p{8cm}}{-LSB- edit -RSB- Fauna \textbf{Bats} were the only mammals of New Zealand until the arrival of humans . 
} & \multicolumn{1}{p{8cm}}{Previously it had been thought that \textbf{bats} were the only terrestrial mammals native to New Zealand .                             
}
\\ \bottomrule
\end{tabular}
\normalsize
\caption{An example of the answers extracted by the $\text{R}^3$ and $\text{SR}^2$ methods, given the question. The words in bold are the extracted answers. The passages are ranked by the highest score (Ranker+Reader) of the answer span in each passage. 
}
\label{tab:analysis2}
\end{table*}

\begin{table}[t]
\centering
\begin{tabular}{lccc}
\toprule
              & TOP-1 & TOP-3 & TOP-5 \\
\midrule
IR            & 19.7 & 36.3 & 44.3 \\
Ranker from $\text{SR}^2$ & 28.8 & 46.4 & 54.9 \\
Ranker from $\text{R}^3$  & 40.3 & 51.3 & 54.5 \\
 \bottomrule
\end{tabular}
\caption{The performance of Rankers (recall of the top-k ranked passages) on the Quasar-T test dataset. This evaluation is simply based on whether the ground-truth appears in the TOP-N passages. IR directly uses the ranking score from raw dataset.}
\label{tab:ranker_analysis}
\end{table}

\subsection{Further Analysis}
In this subsection, we first present an analysis of the improvement of both Ranker and Reader trained with our method, and then discuss ideas for further improvement.\\

\noindent\textbf{Quantitative Analysis}\quad
First, we examine whether our RL approach could help the Ranker overcome the absence of any ground-truth ranking score. To control everything but the change in Ranker, we conduct two experiments combining the same Single Reader with two different Rankers trained from $\text{SR}^2$ and $\text{R}^3$, respectively. Table~\ref{tab:analysis1} shows the results on the Quasar-T test dataset. Note that the Single Reader combined with the Ranker trained from $\text{R}^3$ model achieves an EM 1.3 higher performance than combined with the Ranker from $\text{SR}^2$ which treats all passages containing ground-truth answer as positive cases. That means our proposed Ranker is better than the Ranker normally trained in the distant supervision setting.

We also find that the performance of $\text{R}^3$ can still achieve an EM 1.0 higher than the Single Reader combined with the Ranker from $\text{R}^3$ through Table~\ref{tab:analysis1}. In this setting, the Ranker is the same, while the Reader is trained differently. We infer from this that our proposed methods $\text{R}^3$ can not only improve the Ranker but also the Reader. \\

\noindent\textbf{Potential Improvement}\quad
We offer a statistical analysis to approximate the upper bound achievable by only improving the ranking models.
This is evaluated by computing the QA performance with the best passage among the top-$k$ ranked passages.
Specifically, for each question, we extract one answer from each of the top-$50$ passages retrieved from the IR system, and take the top-$k$ answers with the highest scores according to Eqn.(\ref{eqn:pred_prob}) from these.
Based on the $k$ answer candidates, we compute the \textbf{TOP-k} F1/EM by evaluating on the answer with highest F1/EM score for each question. This is equivalent to having an \emph{oracle ranker} that assigns a $+\infty$ score to the passage (from the passages providing top-k candidates) yielding the best answer candidate.

Table~\ref{tab:analysis_upper} shows a clear gap between TOP-3/5 and TOP-1 QA performances (over 12-20\%). According to our evaluation approach of TOP-k F1/EM and since the same SR model is used, this gap is
solely due to
the oracle ranker. Although our model is far from the oracle performance, it still provides a useful upper bound for improvement. \\

\noindent\textbf{Ranker Performance Analysis}\quad
Next we show the intermediate performance of our method on the ranking step.
Since we do not have the ground-truth for the ranking task, we evaluate on pseudo labels: a passage is considered  positive if it contains the ground-truth answer. Then a ranker's top-$k$ output is considered accurate if any of the $k$ passages contain the answer (i.e. \emph{top-k recall}). 
Note that this way of evaluation on top-1 is consistent with the training objective of the ranker in $\text{SR}^2$.

From the results in Table~\ref{tab:ranker_analysis}, the Ranker from $\text{R}^3$ performs significantly better than the one from $\text{SR}^2$ on top-1 and top-3 performance, despite the fact that it is not directly trained to optimize this pseudo accuracy. Given the evaluation bias that favors the $\text{SR}^2$, this indicates that our $\text{R}^3$ model could make Ranker training easier, compared to training on the objective in Eqn.\ref{eqn:rank_obj_ideal} with pseudo labels.

{Starting from top-5, the Ranker from $\text{R}^3$ gives slightly lower recall. This is because the two Rankers have a similar ability to rank the potentially useful passages in the top-5, but the evaluation bias benefits the $\text{SR}^2$ Ranker. 
Overall, our $\text{R}^3$ could successfully rank the potentially more useful passages to the highest positions (top 1-3), improving the overall QA performance.}

An example in Table~\ref{tab:analysis2} illustrates the importance of ranking.  The passages on the left are from the $\text{R}^3$ Ranker and the ones on the right from the $\text{SR}^2$ Ranker. If $\text{SR}^2$ ranked P2 or P3 higher, it could also have extracted the right answer. In general, if passages that can entail the answer are ranked more accurately, both models could be improved.

\section{Related Work}

\textbf{Open domain question answering} dates back to as early as \cite{green1961baseball} and was popularized with TREC-8 \cite{voorhees1999trec}. The task is to answer a question by exploiting resources such as
documents~\cite{voorhees1999trec}, webpages~\cite{kwok2001scaling,chen2017discriminative} 
or structured knowledge bases~\cite{berant2013semantic,bordes2015large,yu2017improved}.
An early consensus since TREC-8 has produced an approach with three major components: question analysis, document retrieval and ranking, and answer extraction. Although question analysis is relatively mature, answer extraction and document ranking still represent significant challenges.

Very recently, IR plus machine reading comprehension (\textbf{SR-QA}) showed promise for open-domain QA, especially after datasets created specifically for the multiple-passage RC setting \cite{nguyen2016ms,chen2017reading,JoshiTriviaQA2017,dunn2017searchqa,dhingra2017quasar}.
These datasets deal with the end-to-end open-domain QA setting, where  
only question-answer pairs provide supervision.
Similarly to previous work on open-domain QA, existing deep learning based solutions to the above datasets also rely on a document retrieval module to retrieve a list of passages for RC models to extract answers.
Therefore, these approaches suffer from the limitation that the passage ranking scores are determined by n-gram matching (with tf-idf weighting), which is not ideal for QA.

Our ranker module in R$^3$
could help to alleviate the above problem, and
RL is a natural fit to jointly train the ranker and reader since the passages do not have ground-truth labels.
Our work is related to the idea of soft or hard attentions (usually with reinforcement learning) for hierarchical or coarse-to-fine decision sequences making in NLP, where the attentions themselves are latent variables. For example, \citeauthor{lei2016rationalizing}~\citeyear{lei2016rationalizing} propose to first extract informative text fragments then feed them to text classification and question retrieval models.
\citeauthor{cheng2016neural}~\citeyear{cheng2016neural} and \citeauthor{choi2017coarse}~\citeyear{choi2017coarse} proposed coarse-to-fine frameworks with an additional sentence selection step before the original word-level prediction for text summarization and reading comprehension, respectively.
To the best of our knowledge, we are the first apply this kind of framework to the open-domain question answering.

From the method-perspective, our work is most close to \citeauthor{choi2017coarse}~\citeyear{choi2017coarse}'s work in terms of the usage of REINFORCE. Our main aim is to deal with the lack of annotation in the passage selection step, which is a necessary intermediate step in open-domain QA. In comparison, \citeauthor{choi2017coarse}~\citeyear{choi2017coarse} has as its main aim to speed up the RC model in the single passage setting. 
From the motivation-perspective, we are similar to \citeauthor{narasimhan-yala-barzilay:2016:EMNLP2016}~\citeyear{narasimhan-yala-barzilay:2016:EMNLP2016}'s work. Both work aim to find passages easy and suitable for the QA or IE models to extract answers, in order to boost accuracy.

\section{Conclusion}
We have proposed and evaluated $\text{R}^3$, a new open-domain QA framework which combines IR with a deep learning based Ranker and Reader. First the IR model retrieves the top-$N$ passages conditioned on the question. Then the  Ranker and Reader are trained jointly using reinforcement learning to directly optimize the expectation of extracting the ground-truth answer from the retrieved passages.
Our framework achieves the best performance on several QA datasets.

\bibliography{aaai}
\bibliographystyle{aaai}
\end{document}